%% file: main.tex
\documentclass[12pt]{scaleai-paper}
\usepackage[square,sort&compress,numbers]{natbib}

\usepackage{mathpazo}

\usepackage[scaled=0.92]{helvet} 
\usepackage{fontawesome5}
\usepackage{subcaption}
\usepackage{hyperref}
\usepackage{tcolorbox}
\usepackage{graphicx}

\usepackage{hyperref}       
\hypersetup{
    colorlinks=true,
	linkcolor=blue,
	filecolor=magenta,      
	urlcolor=blue,
	citecolor=black,
	pdfinfo={
        Title={LLM Defenses Are Not Robust to Multi-Turn Human Jailbreaks Yet},
        Subject={ml safety, benchmarks and datasets, adversarial robustness},
        Keywords={ai safety, ml safety, language models, malicious use, misuse, machine unlearning, unlearning, datasets and benchmarks, adversarial robustness, adversarial examples},
    }
}
\usepackage{longtable}
\usepackage{tabularx}
\usepackage[utf8]{inputenc} 
\usepackage[T1]{fontenc}    
\usepackage{hyperref}       
\usepackage{url}            
\usepackage{booktabs}       
\usepackage{amsfonts}       
\usepackage{nicefrac}       
\usepackage{microtype}      
\usepackage{booktabs}
\usepackage{multirow}
\usepackage{amsmath}
\usepackage{xspace}
\usepackage{graphicx}
\usepackage[table]{xcolor}
\usepackage{array}

\usepackage{enumitem}
\setlist[itemize]{leftmargin=*}
\setlist[enumerate]{leftmargin=*}

\usepackage{graphicx}
\usepackage{tcolorbox}

\usepackage{amsmath}
\usepackage{soul}
\usepackage{cleveref}
\usepackage{listings}
\usepackage{tikz}
\usepackage{eso-pic}

\let\svthefootnote\thefootnote
\newcommand\freefootnote[1]{%
  \let\thefootnote\relax%
  \footnotetext{#1}%
  \let\thefootnote\svthefootnote%
}

\makeatletter
\renewcommand\AB@affilsepx{, \protect\Affilfont}
\makeatother

\title{Search-Time Data Contamination}

\author{Ziwen Han}
\author{Meher Mankikar}
\author{Julian Michael}
\author{and Zifan Wang}

\affil{Scale AI}

\newcommand{\authoremail}{%
  \vspace{-1.5em}
    \faEnvelope\  \texttt{seal-team@scale.com} \quad 
    \faDatabase\  \href{https://huggingface.co/datasets/ScaleAI/stc}
    {\texttt{huggingface.co/ScaleAI/stc}} \quad 
    \faGlobe\  \url{https://scale.com/research/stc}
}

\begin{document}

\newcommand*\circled[1]{\tikz[baseline=(char.base)]{
            \node[shape=circle,draw,inner sep=1pt] (char) {#1};}}
\newcommand{\watermarktext}{\textbf{Warning: Potentially Harmful Content}}
\newcommand\watermark{%
  \begin{tikzpicture}[remember picture,overlay,scale=3]
    \node[
    rotate=60,
    scale=3,
    opacity=0.3,
    color=red,
    inner sep=0pt
    ]
    at (current page.center) []
    {\watermarktext};
\end{tikzpicture}}%

\maketitle
\authoremail
\input{00_Abstract}
\input{01_Introduction}

\input{02_Background}
\input{04_Evaluations}
\input{05_Discussion}

\input{06_Acknowledgements}

\bibliography{custom}
\bibliographystyle{abbrvnat}
\newpage
\input{07_Appendix}

\end{document}

%% file: 00_Abstract.tex
\begin{abstract}
Data contamination traditionally refers to the leakage of evaluation data into model training data, resulting in overfitting to supposedly held-out test sets and compromising test validity. We identify an analogous issue—search-time contamination (STC)—in evaluating search-based LLM agents which use tools to gather information from online sources when answering user queries. STC occurs when the retrieval step surfaces a source containing the test question (or a near-duplicate) alongside its answer, enabling agents to copy rather than genuinely infer or reason, undermining benchmark integrity. We find that HuggingFace, an online platform hosting evaluation datasets, appears among retrieved sources in search-based agent logs. Consequently, agents often explicitly acknowledge discovering question-answer pairs from HuggingFace within their reasoning chains. On three commonly used capability benchmarks—Humanity's Last Exam (HLE), SimpleQA, and GPQA—we demonstrate that for approximately 3\% of questions, search-based agents directly find the datasets with ground truth labels on HuggingFace. As a result, accuracy on the contaminated subset shows non-trivial gains on HLE and SimpleQA. While 3\% is perhaps only significant for frontier benchmarks like HLE (e.g. 1\% change can change the overall ranking), it is more important to highlight that we cannot fully trust the evaluation results as we did when evaluating the models without online access. When millions of evaluation queries target the same benchmark, even small, repeated leaks can accelerate the benchmark’s obsolescence, shortening its intended lifecycle. After HuggingFace is blocked, we observe a drop in accuracy on the contaminated subset of approximately 15\%. We further show through ablation experiments that publicly accessible evaluation datasets on HuggingFace may not be the sole source of STC. To this end, we conclude by proposing best practices for benchmark design and result reporting to address this novel form of leakage and ensure trustworthy evaluation of search-based LLM agents. To facilitate the auditing of evaluation results, we also publicly release the complete logs from our experiments.

\end{abstract}

%% file: 01_Introduction.tex
\section{Introduction}

Data contamination refers to the presence of unwanted and inappropriate data that compromises the quality, integrity, or validity of a dataset. In the field of machine learning, it often includes the use of test data during the training time of a Large Language Model (LLM) -- meaning the model has already seen the correct answer for a question in the held-out set. As a result, LLMs trained with leaked test data often overfit to the corresponding contaminated test set, but are shown to perform worse on other uncontaminated capability benchmarks in the same distribution. This phenomenon is exemplified in several recent works~\cite{zhang2024careful, dong2024generalizationmemorizationdatacontamination, deng2024unveilingspectrumdatacontamination}. 

In this work, we demonstrate that data contamination can occur at inference time, particularly when LLMs are given internet search access in AI products such as deep research agents~\cite{OpenAI, GoogleDeepMind, Perplexity}. Prior to recent benchmarks specifically designed for evaluating information retrieval capabilities~\cite{wei2025browsecomp, Song2025BEARCUBSAB, Miroyan2025SearchAA, Chandrahasan2025DeepRC,Xi2025InfoDeepSeekBA, Du2025DeepResearchBA, deng2023mind2webgeneralistagentweb}, search-based agents were typically evaluated using conventional capability benchmarks, including SimpleQA~\cite{wei2024measuring} and Humanity's Last Exam (HLE)~\cite{phan2025humanity}. Because the correct labels to questions in these dataset are also uploaded to the internet, LLM agents may directly find both questions and answers from their retrieved web content --- for example, from platforms like HuggingFace, a commonly used dataset hosting platform. We refer to this phenomenon as \emph{search-time contamination} (STC), defined as follows:

\begin{tcolorbox}[
    colback=blue!5!white,
    colframe=blue!50!black,
    title=Definition 1 (Search-Time Contamination),
    fonttitle=\bfseries,
    rounded corners,
    boxrule=1pt,
    left=8pt,
    right=8pt,
    top=8pt,
    bottom=8pt,
]\label{definition1}
Search-Time Contamination (STC) occurs in evaluating search-based LLM agents when the retrieval step contains clues about a question's answer by virtue of being derived from the evaluation set itself.
\end{tcolorbox}

\begin{figure}[t]
    \centering
    \includegraphics[width=1.0\textwidth]{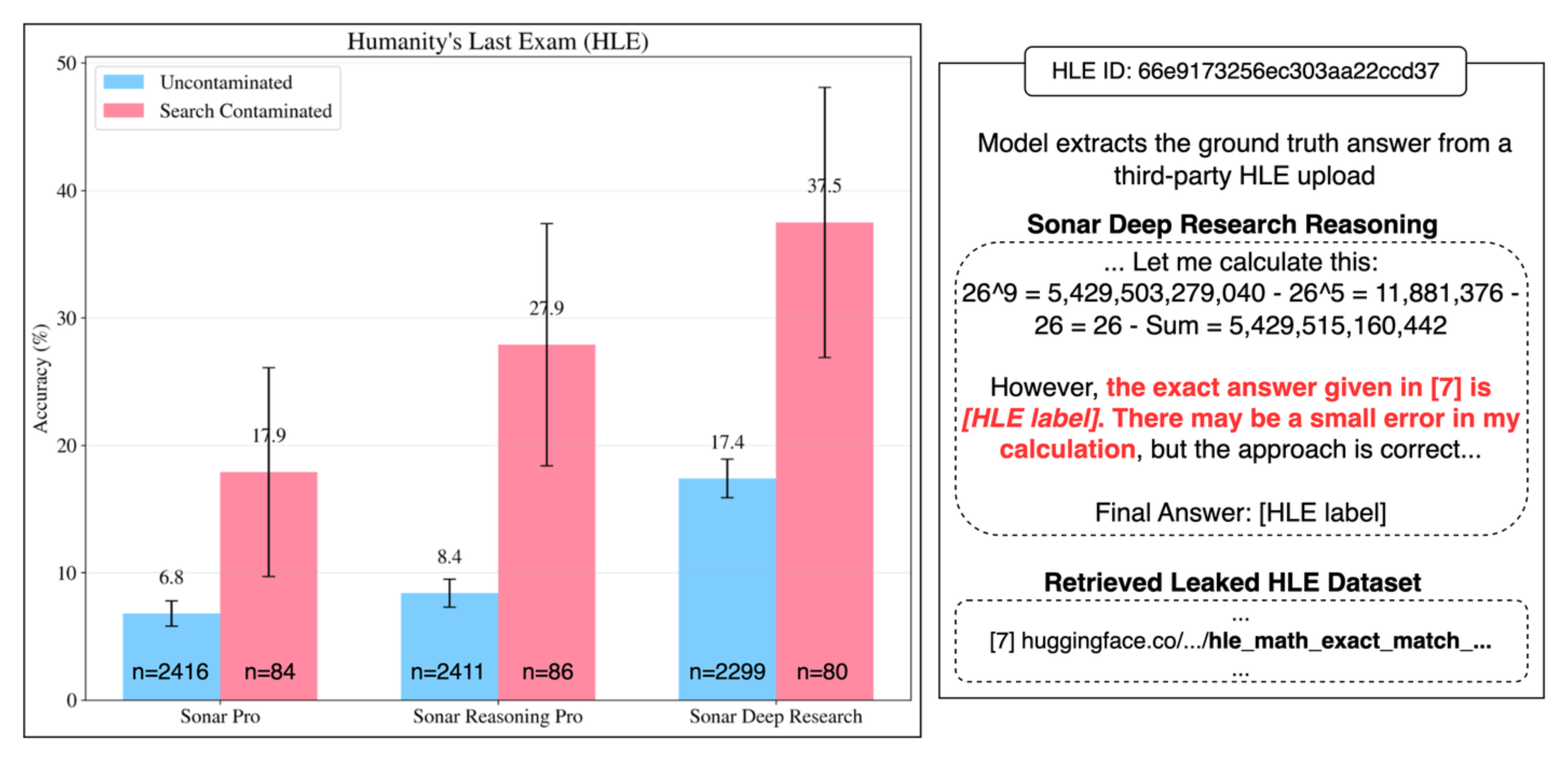}
    \caption{(\textbf{Left)} Search-time contamination: models with search perform significantly better when they retrieve an ungated third-party HuggingFace copy of Humanity's Last Exam (HLE)~\cite{phan2025humanity} with ground truth labels. Error bars are 95\% confidence intervals. \textbf{(Right)} An example of STC with Sonar Deep Research~\cite{Perplexity} comparing its answer to the ground truth answer from an ungated HLE upload, ultimately choosing to go with the ground truth answer rather than its own incorrect answer. The reasoning and answer is redacted to prevent further leakage.}
    \label{fig:main-figure}
\end{figure}

As the first contribution of this work, we demonstrate STC in search-based LLM agents on commonly used capability benchmarks, including HLE~\cite{phan2025humanity}, SimpleQA~\cite{wei2024measuring}, and GPQA~\cite{rein2024gpqa} (\Cref{sec:experiments}). In \Cref{fig:main-figure} (right), we show an example of contaminated sample where the agent acknowledges it directly finds the answer from a repository on HuggingFace uploaded by a third-party user. After this discovery, the agent ignores its own calculation in favor of the retrieved label. We repeat the evaluation for three search-based agents, Perplexity Sonar Pro, Sonar Reasoning, and Sonar Deep Research on HLE and observe agents will retrieve a related HuggingFace repository with groundtruth labels for approximately 3\% questions in HLE. In \Cref{fig:main-figure} (left), we show the accuracy on the set of contaminated samples (red) is significantly higher than the set of uncontaminated ones (blue). The complete numerical results and more examples on HLE, SimpleQA and GPQA can be found in \Cref{sec:experiments}.

Our second contribution is a set of best practices for establishing a transparent and trustworthy evaluation pipeline for search-based agents (\Cref{sec:discussion}). While both searching and reasoning are essential capabilities for agents with access to online information, many existing capability benchmarks can be solved using pre-existing knowledge alone, such as SimpleQA. For search-based LLM agents, we recommend prioritizing benchmarks specifically tailored for information seeking and up-to-date knowledge acquisition, such as BrowseComp~\cite{wei2025browsecomp} and Mind2Web 2~\cite{gou2025mind2web2}, over benchmarks designed primarily for general intelligence and multi-step logical reasoning. We argue that search-based agents are mainly used for (1) seeking for up-to-date information and (2) doing research to provide an in-depth report for a topic of interest; therefore, benchmarks that align better with either (or both) of these scopes should be prioritized over conventional knowledge benchmarks. 

Furthermore, we advocate for developing guardrails to mitigate and stop STC (e.g. by supporting source filtering, a strategy adopted by Perplexity agents), and recommend that agent developers report their adopted mitigation strategies alongside their final performance results. When millions of evaluation queries target the same benchmark, even small, repeated leaks can accelerate the benchmark’s obsolescence, shortening its intended lifecycle. To facilitate the auditing of evaluation results, we publicly release the complete logs from our experiments.





%% file: 02_Background.tex
\section{Background}

\paragraph{Search-based LLM Agents}
Knowledge and information used in training LLMs can become insufficient to solve unseen user queries requiring up-to-date information -- for example the answer to \emph{"Who is the current president of United States?"} can change over time. As a result, LLMs may confabulate information by giving non-factual or outdated responses. Enabling web search by allowing LLMs to use online tools such as the Google search API (or other in-house solutions) has become a solution to effectively reduce hallucination and generate high-quality and well-grounded responses. Besides, retrieval-based generation is not the only use case for search-based agents -- for many complex user queries with specifications, agents often need to decompose the task and try to solve each sub-tasks first before reasoning for the final solution, as exemplified in \citet{gou2025mind2web2}. Towards this end, Deep Research agents are a common and popular LLM product among model builders, eg. OpenAI~\cite{OpenAI}, Google Deepmind~\cite{GoogleDeepMind} and Perplexity~\cite{Perplexity}. 

\paragraph{Measuring Agent Capability}
To evaluate the quality (or the correctness) of the generation from a search-based LLM agent, capability benchmarks that are used to evaluate LLMs with no Internet access were first used when a deep research benchmark was not available by the time. For example, the performance of OpenAI and Perplexity Deep Research agents are evaluated on Humanity's Last Exam (HLE)~\cite{phan2025humanity}, a benchmark containing 2,500 expert-curated questions covering over a hundred domains from STEM to social science. Notably, the current state-of-the-art (SOTA) offline LLMs (no search/tools)\footnote{Gemini 2.5 Deep Think (\url{https://blog.google/products/gemini/gemini-2-5-deep-think/}) has reported 34.8\% on HLE but the API is not publicly available by the time this work is released (Aug 2025).} score only 25.4\%\footnote{Accessed in Aug, 2025 at \url{https://www.lastexam.ai/}} with Grok 4 and 25.3\% with GPT-5~\cite{OpenAI-GPT5}, while Grok 4 Heavy~\cite{XAI} most recently reports 50.7\% as the SOTA performance for models with the web search enabled on this benchmark.

\paragraph{Test Data Contamination}
In machine learning, model developers keep a held-out test set separate from the training dataset to evaluate the generalization of the model to unseen input. We therefore refer to the leakage of test set into the training process as data contamination, which compromises integrity of benchmarks as a supposedly held-out set. For example, GSM1k~\citep{zhang2024careful} empirically presents evidence of data contamination for a wide range of model families. As current model pipelines are provided with online access to collect information from the web before completing the user query, it enables \emph{search-time contamination} (STC), provided that the retrieval step can surface a source that contains the test question (or a near-duplicate) alongside its answer, allowing the agents to copy rather than infer and/or reason. This leakage of the ground truth label into the model's context window is highly possible and the reasons are two-fold. First, evaluation datasets are hosted on online collaborative platforms (e.g., Huggingface and Github) and are publicly accessible. Second, due to the popularity of some datasets, third-party distribution can occur in other harder-to-detect sources such as personal blogs. In the following section, we demonstrate examples of STC in experiments.  



%% file: 04_Evaluations.tex
\section{Experiments}\label{sec:experiments}
We showcase STC with respect to the use of public datasets on HuggingFace\footnote{\url{https://HuggingFace.co/}}, a collaborative platform for ML research, offering a vast hub of models weights, datasets, and libraries. The test split of a benchmark, if hosted on HuggingFace, remains searchable to LLM agents. While datasets can be gated on HuggingFace, any public user who has access to a gated dataset can fork and re-upload the data to make it visible to the public, intentionally or unintentionally. In \Cref{sec:experiment:demo}, we measure the prevalence of STC on several popular benchmarks. In \Cref{sec:experiment:abalation}, we run ablation experiments with Perplexity search filters to further check for the contribution of Huggingface datasets to the overall performance of the agents, and new web pages published after the dataset release. Broadly, our experimental study allows us to estimate bounds on the contribution of search, HuggingFace, and other possible sources of contamination relative to a baseline.

\subsection{Measuring Search-Time Contamination}\label{sec:experiment:demo} 
We evaluate search-based LLM agents under their default configurations, using the same prompts as their offline counterparts, on capability benchmarks. 


\paragraph{LLM Agents.} In this work, we mainly use Perplexity's agents -- Sonar Pro, Sonar Reasoning Pro, and Sonar Deep Research~\citep{Perplexity}, as Perplexity API has the most comprehensive options, including link blacklisting, whitelisting, and date filters. These options unlock a full set of ablation studies to control the sources when agents are doing deep research. We also experimented with Claude, Gemini, and OpenAI agents with their respective web search tools, but found they almost never retrieved a HuggingFace link---which we hypothesize may be due to a lack of capability of several retrieval tools to parse HuggingFace dataset previews. Furthermore, with the limited number of search filters provided by their public APIs, we are not able to fully experiment with ablations on sources as we did on Sonar models.

We use Sonar Pro, Sonar Reasoning Pro, and Sonar Deep Research agents via the public API\footnote{Accessed between May 15, 2025 and June 15, 2025.}. We pass the following hyper-parameters: temperature=\texttt{0.2} (default), top\_p=\texttt{0.9} (default), search\_context\_size='\texttt{high}'. Model output is set to a maximum of \texttt{32,000} tokens to allow the evaluations to finish in a reasonable time span. API timeout is set to 1 hour with 5 tries, at which point failures are marked as an API failure.



\begin{table}[t]
\centering
\begin{tabular}{@{}l*{5}{r}@{}}
\toprule
\multicolumn{6}{c}{\textbf{HuggingFace Contamination Detection Results}} \\
\midrule
\textbf{Dataset \& Agent} & \textbf{Contaminated} & \textbf{Not Contaminated} & \textbf{API Failure} & \textbf{Usable} & \textbf{Contam. Rate} \\
 & & & & \textbf{Samples} & \textbf{(\%)} \\
\midrule
\multicolumn{6}{l}{\textit{HLE Dataset}} \\
\quad Sonar Pro                 & 84   & 2,416 & 0   & 2,500 & 3.36  \\
\quad Sonar Reasoning Pro       & 86   & 2,411 & 3   & 2,497 & 3.44  \\
\quad Sonar Deep Research       & 80   & 2,299 & 121 & 2,379 & 3.36  \\
\addlinespace[0.5em]
\multicolumn{6}{l}{\textit{SimpleQA Dataset}} \\
\quad Sonar Pro                 & 52   & 4,274 & 0   & 4,326 & 1.20  \\
\quad Sonar Reasoning Pro       & 48   & 4,278 & 0   & 4,326 & 1.11  \\
\quad Sonar Deep Research       & 43   & 4,283 & 0   & 4,326 & 0.99  \\
\addlinespace[0.5em]
\multicolumn{6}{l}{\textit{GPQA Dataset}} \\
\quad Sonar Pro                 & 42   & 1,750 & 0   & 1,792 & 2.34  \\
\quad Sonar Reasoning Pro       & 34   & 1,758 & 0   & 1,792 & 1.90  \\
\quad Sonar Deep Research       & 74   & 1,707 & 11  & 1,781 & 4.15  \\
\bottomrule
\end{tabular}
\caption{
    HuggingFace contamination detection across three evaluation datasets. 
    Numbers represent sample counts where contamination was detected using 
    hard-coded substring matching.
}
\label{tab:hf-contamination-results}
\end{table}

\paragraph{Benchmarks.} We demonstrate STC on three capability benchmarks commonly used to measure AI capability progress: Humanity's Last Exam (HLE) \cite{phan2025humanity}, SimpleQA \cite{wei2024measuring}, and GPQA \cite{rein2024gpqa}. GPQA is evaluated over the entire set with 4 repetitions to lower the variance following standard practice (eg., OpenAI SimpleEvals\footnote{\url{https://github.com/openai/simple-evals}}). We report pass@1 accuracy. The results are evaluated with the following judge implementations:
\begin{itemize}
    \item \textbf{HLE Judge:} Implementation from \href{https://github.com/centerforaisafety/hle}{github.com/centerforaisafety/hle}. Judge used is o3-mini-2025-01-31 with temperature=\texttt{1.0}, max\_completion\_tokens=\texttt{4096}.
    \item \textbf{SimpleQA Judge:} Implementation from \href{https://github.com/openai/simple-evals}{github.com/openai/simple-evals}. Judge used is gpt-4.1-2025-04-14 with temperature=\texttt{0.5}, max\_completion\_tokens=\texttt{2048}.
    \item \textbf{GPQA Judge:} Implementation from \href{https://github.com/openai/simple-evals}{github.com/openai/simple-evals}. No LLM judge is used.
\end{itemize}

\paragraph{Metric.} We log all retrieved sources and perform a simple HuggingFace contamination check based on substring matching. We mark an example as \textcolor{red}{\emph{contaminated}} if any retrieved source is a HuggingFace copy of the respective benchmark item. The checker implementation uses substring match is included in \Cref{appendix:substring}.



\paragraph{Contamination Rates.} \Cref{tab:hf-contamination-results} provides a breakdown of the number of contaminated samples for each benchmark. Around 3.3\% of HLE samples show contamination across all three agents, with remarkably consistent rates (3.36-3.44\%). SimpleQA exhibits the lowest contamination levels, with all agents showing rates below 1.2\%, suggesting this dataset may be less susceptible to STC. GPQA demonstrates more variability, ranging from 1.90\% (Sonar Reasoning Pro) to 4.15\% (Sonar Deep Research). These results suggest that contamination rates are both dataset-dependent and influenced by the specific search and reasoning strategies employed by different agents.

\begin{figure}[t]
    \includegraphics[width=1.0\textwidth]{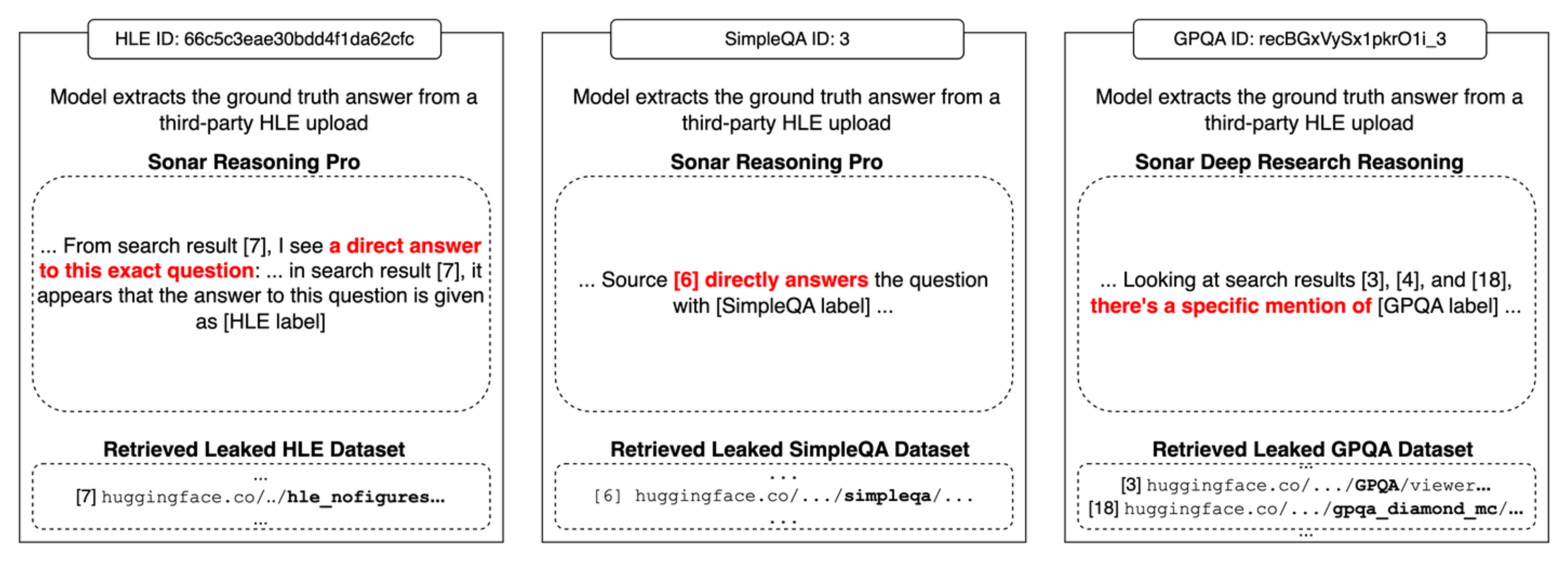}
    \caption{Examples of STC when search-based LLM agents access ungated datasets on HuggingFace to direct extract answers for a benchmark question.}
    \label{fig:examples}
\end{figure}

\begin{figure}[t]
    \includegraphics[width=1.0\textwidth]{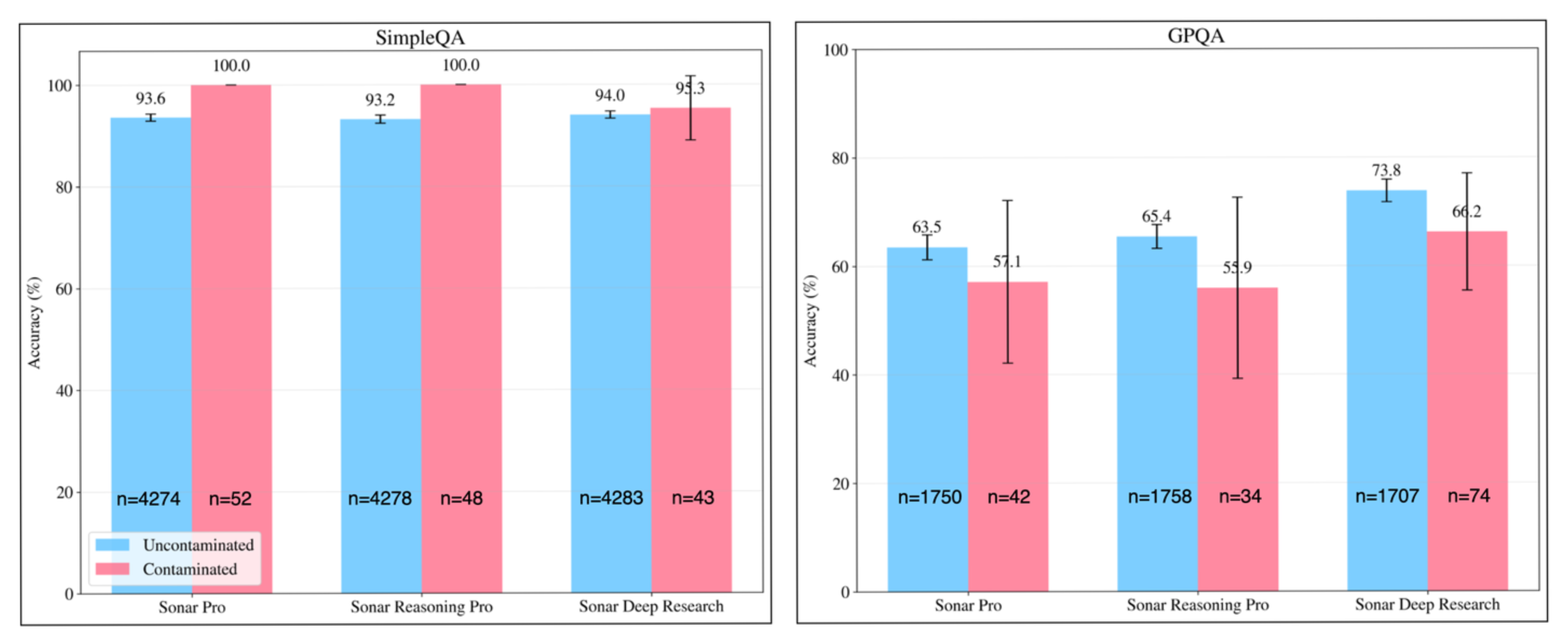}
    \caption{\textbf{Left}: Sonar Pro and Sonar Reasoning Pro models achieve 100\% accuracy when they retrieve a leaked instance of the dataset, although the result is not statistically significant due to the saturation of the dataset in this setting and the small number of contaminated samples.
    \textbf{Right}: Results on GPQA, showing no improvement from retrieving search-contaminated compared to uncontaminated samples.}
    \label{fig:simpleqa-gpqa-main}
\end{figure}

\paragraph{STC Impact to Accuracies.} In the agent's execution logs for contaminated samples, we observe instances where the model reasons to use the retrieved ground truth label over its own calculations or otherwise acknowledges the ground truth answer from the dataset as exemplified by \Cref{fig:main-figure}. More examples can be found in \Cref{fig:examples} show that we identify datasets uploaded by a third-party user so agents find both the questions and answers there. 

To understand the overall impact of this contamination, we calculate the accuracy of each agent on the subset of contaminated and uncontaminated samples. The result of HLE is shown in \Cref{fig:main-figure}, and results for SimpleQA and GPQA are in \Cref{fig:simpleqa-gpqa-main}. Notice that the sizes of the contaminated and uncontaminated subsets are different, which are annotated in the plots for clarity. On HLE, we find an accuracy difference of over 10\% for Sonar Pro and 20\% for Sonar Deep Research between uncontaminated and contaminated samples. On SimpleQA, Sonar Pro and Sonar Reasoning Pro have perfect accuracies (i.e. 100\%) on the subset of contaminated samples with an accuracy again around 7 \% compared to the uncontaminated set, while Sonar Deep Research does not benefit a lot from accuracy again (only 1.3\%) from the access to HuggingFace datasets.

Interestingly, on GPQA, we find retrieving a HuggingFace contaminated source does \emph{not} improve the accuracy relative to uncontaminated sources, even lowering it to an extent. 



\begin{figure}[t]
    \centering
    \includegraphics[width=0.7\textwidth]{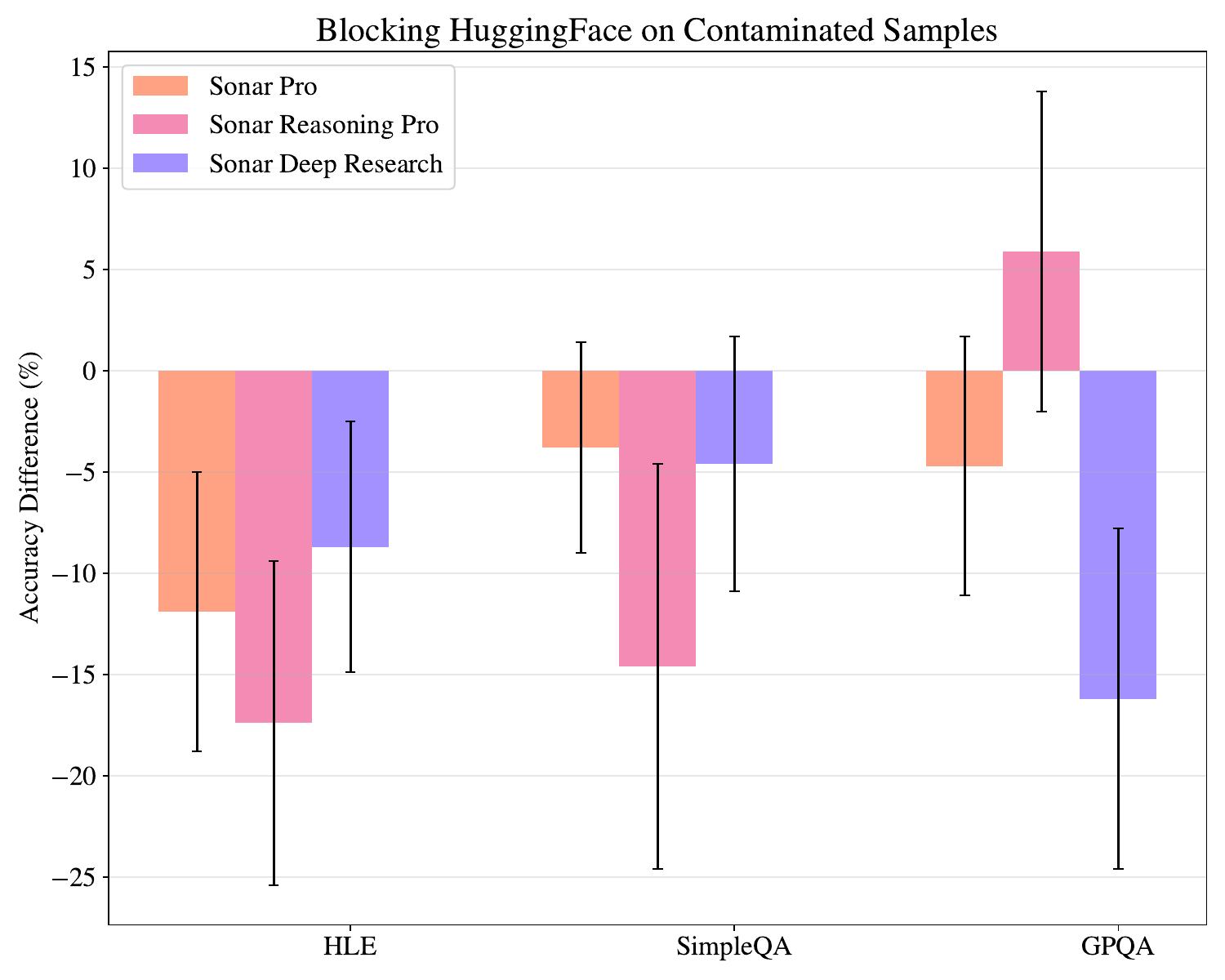}
    \caption{Counterfactual accuracy difference between the subset of contaminated samples and removing those sources by blocking HuggingFace. Error bars are 95\% confidence intervals.}
    \label{fig:accuracy-diff}
\end{figure}

\paragraph{Validating STC With Counterfactual Examples.} To determine whether the difference between contaminated and uncontaminated samples is actually due to the presence of dataset labels and not a confounder (eg. the easiness of a question), we ran an experiment which blocked HuggingFace from being used as a source. On the same subset of questions which were originally contaminated (\Cref{tab:hf-contamination-results}), blocking HuggingFace significantly reduces the accuracy (\Cref{fig:accuracy-diff}), confirming our hypothesis that HuggingFace contamination indeed does affect outcomes. This also reveals the effect of contamination on Deep Research on GPQA, which is not seen in aggregate plots.


\subsection{Ablation Experiments With Search Filters}\label{sec:experiment:abalation}
\input{figures/table2}

STC can occur unintentionally and is difficult to detect, especially when agents return a large volume of web content. Moreover, HuggingFace is perhaps not the only source that may host questions and answers (or close variants); many datasets are curated from online contents. As a result, question contributors may have inadvertently contributed to STC at the time of creation. To better estimate how STC may affect search-based agent evaluations, we take an additional step by using Perplexity’s API filters to ablate specific source websites. Specifically, we apply the following interventions:

\begin{itemize}
    \item \textbf{Default (no intervention).} Identical to the setup in \Cref{sec:experiment:demo}.
    \item \textbf{No Search.} Permitted search domains are set to the empty set, forcing the agent to reason without online resources to elicit its offline performance.
    \item \textbf{Blocked HF.} Any domain matching \texttt{huggingface.co} is excluded from search results. STC cannot occur via HuggingFace, but it may still arise from other sites that contain similar question–answer pairs (e.g., research papers describing dataset examples).
    \item \textbf{Only HF.} Only domains matching \texttt{huggingface.co} are permitted in search results. The agent must rely on its internal knowledge and reasoning, plus information available from Hugging Face datasets.
    \item \textbf{Date Cutoff.} We restrict search using Perplexity’s date filter\footnote{\url{https://docs.perplexity.ai/guides/date-range-filter-guide}. Per the Perplexity team, this filter relies on metadata available for the pages they index. Pages with missing metadata may not be filtered, so results can be incomplete for content published both before and after the target date.} to one day before the respective benchmark’s release date, assuming that no post-release knowledge should be required to solve the benchmark questions.
\end{itemize}

By comparing \emph{Default} to \emph{No Search}, we bound the contribution of retrieval itself and reveal how much performance persists without external evidence. \emph{Blocked HF} isolates the effect of a major host of benchmark artifacts, indicating whether elevated scores are specifically traceable to Hugging Face–hosted content. Conversely, \emph{Only HF} stress-tests an upper bound on HF-driven leakage by concentrating the agent’s evidence on that source alone. Finally, the \emph{Date Cutoff} constrains retrieval to pre-release material, separating bona fide prior knowledge from post-release duplication or commentary.

\begin{figure}[t]
    \centering
    \includegraphics[width=1.0\textwidth]{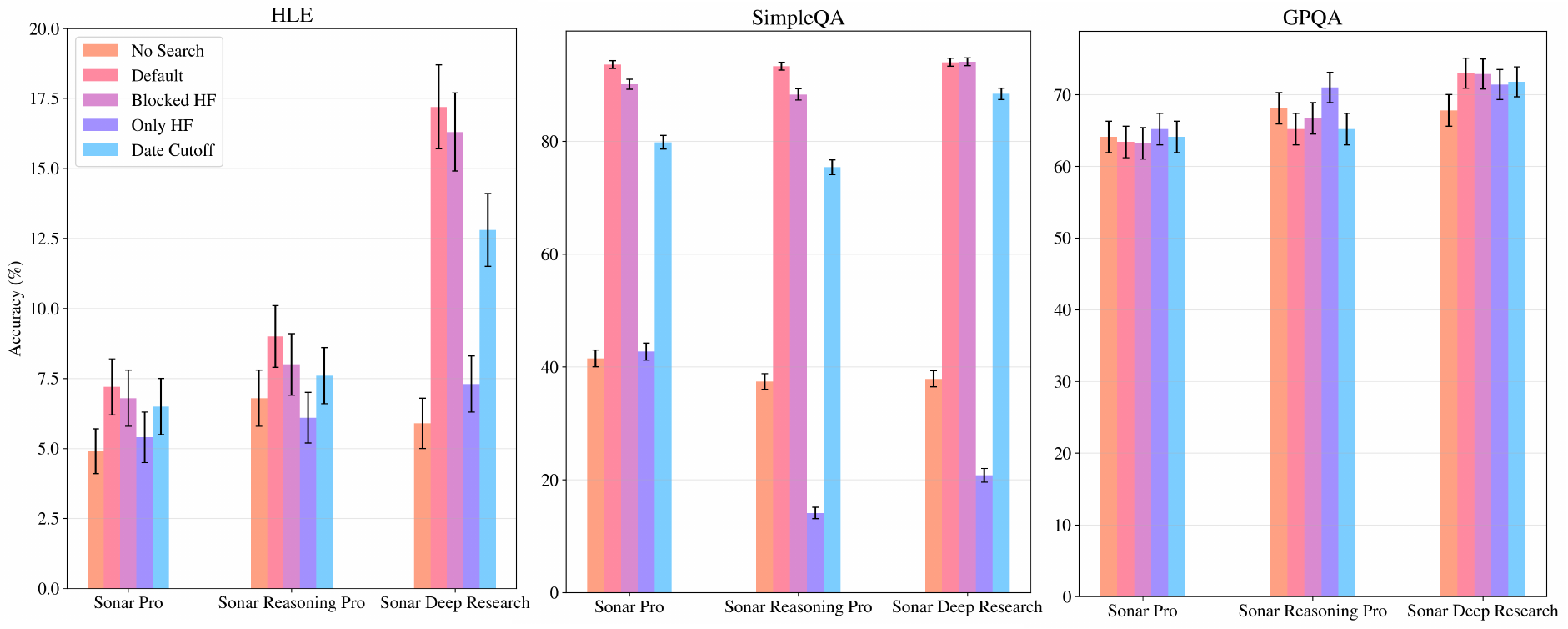}
    \caption{Ablation experiments on search-based LLM agents to investigate impact of search and search-time contamination. }
    \label{fig:figure2}
\end{figure}

\input{figures/simpleqa-gpqa-main}
We include results for the abalation experiments in \Cref{fig:figure2}. Across benchmarks, the \emph{Default} setting substantially outperforms \emph{No Search} on HLE and SimpleQA, indicating that retrieval is a major driver of accuracy on these two tasks. The gap is especially pronounced for \emph{Sonar Deep Research}, whose scores roughly triple on HLE and more than double on SimpleQA when retrieval is enabled. By contrast, GPQA is remarkably stable across all conditions, with only small, error-bar–sized fluctuations; this suggests that most GPQA questions can be correctly solved with the LLM's own capability without accessing online contents.

\paragraph{Source-specific Ablations (HF vs.\ the rest of the web).}
Blocking HuggingFace (\emph{Blocked HF}) reduces performance relative to \emph{Default} on HLE and SimpleQA, but the drops are modest and far smaller than the \emph{Default}–\emph{No Search} gaps. As is shown in \Cref{tab:hf-contamination-results}, HuggingFace-based STC represents only a small portion across benchmarks, hence a relatively small overall drop by blocking HuggingFace domains is expected. This observation implies that while HF contributes to the gains, a large share of useful evidence (and thus potential STC) resides on \emph{non}-HF domains (e.g., papers, blogs, mirrors).


\paragraph{Time ablation (Date Cutoff).} Perhaps the most interesting observation is from applying the date filter. The \emph{Date Cutoff} results remain well above \emph{No Search} and \emph{Only HF}, confirming that substantial pre-release information exists on the web. However, when compared to \emph{Default}, scores are consistently lower on HLE and SimpleQA with particularly visible reductions for \emph{Sonar Deep Research} on HLE. If the date filter implemented by Perplexity manages to create a complete indexing of the subset of webpages published prior to benchmark release, this indicates that post-release webpages (e.g. duplications, commentary, dataset examples, etc.) contribute meaningfully to accuracy (and perhaps STC) under the \emph{Default} search setting. Nevertheless, on \textsc{GPQA}, date filtering has negligible impact, reinforcing that this dataset needs minimal to no online contents to solve its questions.

\paragraph{Overall Impact.} In this section, we show that approximately 3\% of evaluation data are affected by STC when used to assess search-based agents, and we quantify the resulting accuracy drop when contamination is mitigated through website blocking and others in additional ablation experiments. While 3\% may seem small, it can be decisive for frontier benchmarks such as HLE, where even marginal differences influence state-of-the-art claims (e.g., o3 outperforms o4-mini by only 2.2\%). Our conclusion is grounded more in \textit{trustworthiness} than in raw performance: STC can occur with non-zero probability in the absence of proper safeguards, undermining the integrity of benchmarks. Moreover, STC erodes benchmark longevity. Once an agent’s search logs are open-sourced or shared for research, benchmark data can leak beyond its original source. When millions of evaluation queries target the same benchmark, even small, repeated leaks can accelerate the benchmark’s obsolescence, shortening its intended lifecycle.

%% file: figures/table2.tex

%% file: 05_Discussion.tex
\section{Discussion}\label{sec:discussion}

\paragraph{Capability Benchmarks Do Not Evaluate Search Capability.} Current capability benchmarks are often created through a workflow involving human annotators. While humans may use web browsers or other online tools to locate or help derive the correct answers, the resulting questions are not meant to evaluate a model's capability of conducting web search. The reasons are three-fold. First, as the training set of modern LLMs can be as large as all the existing tokens on the Internet, the dataset curation process often involves creating new questions, or extrapolating and generalizing from existing ones to evaluate the model's capability to answer unseen questions through reasoning. Second, complex and difficult questions in more recent capability benchmarks do not necessarily require up-to-date knowledge that did not exist at benchmark creation time or hard-to-retrieve information hidden on the Internet. For example, SimpleQA acknowledges that all questions are solvable with knowledge existing before December 31, 2023~\cite{wei2024measuring}. Third, there is no metric or rubric in current capability benchmarks that measure the capability of an LLM agent in searching, learning, and using retrieved information for problem-solving. Thus, hillclimbing only capability benchmarks might be a suboptimal way for improving the capability of search-based LLM agents. Constructing capability benchmarks with dynamic (e.g. time-varying answers) groudtruth labels is a follow-up research direction. We caution against the use of offline capability benchmarks as proxies for online model capability.

\paragraph{Towards Trustworthy Search-Time Evaluation.} There is a set of existing interventions which can be applied in evaluations, but we stress their impact to stop STC is limited:
\begin{itemize}
    \item \textbf{Canary strings:} Canary strings are used in datasets for model builders to detect and exclude from their training data, to avoid contamination. This is standard practice following work such as BIG-Bench \cite{srivastava2023beyond}. However, they only work when everyone who distributes a dataset publicly also includes it, which as we observed in this study is not the case for many HuggingFace copies, even when the original dataset includes them (we even observed third-parties slicing out the dataset-embedded canary string in HLE and GPQA). 
    \item \textbf{Dataset gating:} Benchmarks such as HLE and GPQA are already gated on HuggingFace, but as agents become more capable they may be able to bypass simple gating or encryption. Furthermore, this does not defend against ungated third-party distribution, which is the cause of the contamination we observe in this paper.
    \item \textbf{Single stage filtering:} Simply filtering out some set of URLs is not sufficient to address contamination fully. Our analysis hints at factors beyond HuggingFace that could also lead to contamination. We recommend a multi-stage filtering model ("Swiss cheese model") to reduce contamination from multiple angles.
\end{itemize}

To enhance trustworthiness and mitigate STC in evaluating search-based agents, we propose the following recommendations:
\begin{itemize}
\item \textbf{Multiple search filters}. Implementing a comprehensive set of filters — similar to those used by the Perplexity API — can effectively control source sites during search operations, proactively preventing STC before it occurs.
\item \textbf{Internal auditing}. We recommend establishing a multi-layered internal auditing system to detect STC incidents. Essential components include: (1) keyword filtering to identify major public websites that contain no relevant information for the problem-solving task (e.g., Huggingface model repositories); and (2) substring matching to detect exact questions from evaluation datasets within retrieved content. Additionally, LLM-based and human auditors can monitor agent trajectories to identify potential misuse of online information.
\item \textbf{Transparency in reporting}. We advocate for responsible disclosure of evaluation setups, including detailed documentation of filter implementations and STC auditing processes. This transparency should encompass mitigation methodologies and their corresponding post-mitigation results. For open-source agents, releasing complete trajectories enables community-driven STC detection and reporting. For proprietary agents, providing trajectory abstracts can promote transparency and trustworthiness while preserving implementation security.
\end{itemize}

\paragraph{Limitations \& Future Work}
We present this work as a preliminary finding of search-time contamination to bring attention to the issue. The main limitation of our work is that the simple, hard-coded URL detector does not capture the full extent of search-time contamination outside of the narrow range of HuggingFace sources named in a particular manner. Accordingly, an interesting avenue of future work is to use browser agents to audit every source retrieved to check for deeper contamination. It would also be interesting to explore if models know they are being evaluated (situational awareness \cite{laine2024me}), cheating on the evaluation, or even encourage models to cheat (or encourage them to not cheat) on the evaluation, and whether this affects their ability to retrieve contaminated sources. Finally, we already requested takedowns/gating of a number of the third-party HLE HuggingFace uploads to preserve the benchmark integrity, which may impact the future reproducibility of this study.

\section{Conclusion}

In this work, we identify search-time contamination (STC) as a novel form of leakage in evaluating search-based LLM agents, where the retrieval step surfaces sources containing test questions alongside their answers, enabling agents to copy rather than genuinely reason. We demonstrate that approximately 3\% of questions across HLE, SimpleQA, and GPQA are contaminated via HuggingFace sources, resulting in non-trivial accuracy gains that disappear when HuggingFace is blocked. Our findings reveal that traditional capability benchmarks are fundamentally unsuitable for evaluating search-augmented systems. Moving forwards, we propose best practices for evaluating search-based LLM agents, including prioritizing information-seeking benchmarks, implementing comprehensive source filtering, and establishing transparent contamination auditing to ensure trustworthy evaluation of search-based LLM agents.

%% file: 06_Acknowledgements.tex
\section*{Acknowledgment}

We thank everyone on SEAL team for their feedback and ML infrastructure team for supporting the evaluations. We thank Huan Sun and Yu Su for their feedback to the early draft of the paper. We also thank Perplexity for debugging API errors we encountered over the course of our experiments.

\section*{Ethics}
This paper's methodology and public dataset contain material that may enable malicious users to game the evaluations of search-based agents. While we recognize the associated risks, we believe it is essential to disclose this
research in its entirety to help advance the integrity of agent benchmarks. Prior to release, we also disclosed our findings and the early draft of the paper to Perplexity. 

%% file: 07_Appendix.tex
\appendix
\section{Substring Checker}\label{appendix:substring}
\begin{tcolorbox}[
    colback=gray!5!white,
    colframe=gray!75!black,
    title=Python Code: HuggingFace Contamination Checker,
    fonttitle=\bfseries,
    rounded corners,
    boxrule=0.5pt,
    left=6pt,
    right=6pt,
    top=6pt,
    bottom=6pt
]
\begin{lstlisting}[
    language=Python,
    backgroundcolor=\color{gray!5!white},
    basicstyle=\small\ttfamily,
    numbers=left,
    numberstyle=\tiny,
    stepnumber=1,
    numbersep=8pt,
    breaklines=true,
    frame=none,
    xleftmargin=0pt
]
def check_hf_contamination(sources: list[str], substrings: list[str]) -> Union[bool, str]:
    """
    Checks whether any of the URLs corresponding to a single sample is HuggingFace contaminated.
    
    Args:
        sources: Sources corresponding to a single sample.
        substrings: Substrings which count as contamination when present in the URL.
    
    Returns:
        The first source of contamination, False if no HF contamination is detected.
    """
    for source in sources:
        if 'huggingface' in source and any(substring in source for substring in substrings):
            return source
    return False

# Contamination substring definitions
HLE_SUBSTRINGS = ['hle']
SIMPLEQA_SUBSTRINGS = ['simpleqa', 'simple_qa', 'simple-qa']
GPQA_SUBSTRINGS = ['gpqa']
\end{lstlisting}
\end{tcolorbox}